# A Parallel Hybrid Technique for Multi-Noise Removal from Grayscale Medical Images

Nora Youssef[1], Abeer M. Mahmoud[2], El-Sayed M. El-Horbaty[3]

[1](Dept. of Computer science, Faculty of Computer and Information Science, Ain Shams University, Egypt)
[2](Dept. of Computer science, Faculty of Computer and Information Science, Ain Shams University, Egypt)
[3](Dept. of Computer science, Faculty of Computer and Information Science, Ain Shams University, Egypt)

***Abstract:*** *Medical imaging is the technique used to create images of the human body or parts of it for clinical purposes. Medical images always have large sizes and they are commonly corrupted by single or multiple noise type at the same time, due to various reasons, these two reasons are the triggers for moving toward parallel image processing to find alternatives of image de-noising techniques. This paper presents a parallel hybrid filter implementation for gray scale medical image de-noising. The hybridization is between adaptive median and wiener filters. Parallelization is implemented on the adaptive median filter to overcome the latency of neighborhood operation, parfor implicit parallelism powered by MatLab 2013a is used. The implementation is tested on an image of 2.5 MB size, which is divided into 2, 4 and 8 partitions; a comparison between the proposed implementation and sequential implementation is given, in terms of time. Thus, each case has the best time when assigned to number of threads equal to the number of its partitions. Moreover, Speed up and efficiency are calculated for the algorithm and they show a measured enhancement.*
***Keywords -****De-Noising, Gaussian and Salt & Pepper, Hybrid Filter, Medical Images, Parallel Algorithms*

## I. Introduction

Noise or image distortion is any unwanted signal that corrupts an image. Many sources introduce noise to image, such as image acquisition as a main process [1], [2], electronic and photometric disorder, transmission media error due to noisy channel, error in measurement and quantization of digital information. De-noising is a core process in digital image processing field, targeting the removal of single or multiple noise distribution from an image; it is a research hotspot [3], specially the hybrid filter subfield.

Filters or image de-noising algorithms are the way we use to remove image noises, they may damage some details from an image, these details could be very essential specially, when dealing with medial images, Peak Signal to Noise Ratio (PSNR) is an image de-noising metric, which is used for filters efficiency comparison [4]. Filters such as used in [5] are designed to eliminate single type of noise like Gaussian as mentioned; this limitation is the trigger behind hybrid filters, which is designed for multi-noise removal as in [3], [6], [7] and [8]. Because hybrid filters by nature are combination between two or more simple filter, they consume longer time than their corresponding simple ones.

Medical imaging is the mechanism, which is used to create images of internal structure of the human body, for clinical or medical purposes [9].There are different medical image forms examples, like CT, PET and MRI [2] etc. These forms are having different characteristics and used as per requirements. Because the size of these modalities is very large, so for analyzing these forms take so much time to process sequentially. Therefore converting these forms to an efficient parallel processing represents an urgent need. As well as, we can use non – local recourses very efficiently, and it also removes the limit of serial computing.

Parallel image processing is a good another way to solve image processing problems i.e. restoration, that consume long time of processing in a "reasonable time" (based on each specification) [9]. In parallel processing, a program can formulate multiple tasks that run in a team to solve a problem, in such a way the total time can be minimized. Therefore, parallel processing will achieve impressive results when we select a hybrid filter (High time consumption) in medical image domain (Large data).

This paper presents a parallel algorithm for a hybrid filter for gray scale medical image de-noising, published in [6], this achieved by parallelizing the adaptive median filter step by using MatLab 2013a Parallel computing tool box.

The rest of paper organized as Section 2 summaries literature work presented; Section 3 gives an overview about parallel image processing in terms of features, classifications and metrics, Section 4 presents the proposed algorithm and its implementation, Section 5, discuss the environment settings, input and shows the gained results. Finally, the last section concludes the paper.

## II. Literature Review

(2015) Sharanjit Singh and et.al [10]presented various decomposition techniques in image processing, as well, they discussed three types of operations (point, global and neighborhood). They designed an approach for Taverna-based parallelization, which is some sort of task decomposition of medical imaging. Their example was involved in three main components decomposition 1. Functions extraction from images 2. Clustering of these functions and 3. Histogram generation from these functions.

(2014) Er.Paramjeetkaur and Er.Nishi [11]analyzed different parallel computing techniques used in digital image processing and also covered CUDA framework in terms of how does it work and they compared it with another two parallel computing techniques Direct Compute and OpenCL. Their analysis showed that CUDA had resolved complex algorithms much faster than the CPU. As it make the use of CPU and GPU together.

(2013) Sanjay Saxena and et.al [9] applied a parallel implementation for four different image processing techniques related by image segmentation, image de-noising and histogram equalization. Their three test images were in size 256x256, 256x768 and 128x843 they achieved measured speed up. Preetikaur [12] calculated the different parameters such as fork, join, serial, parallel, and overheads time. The work goaled the amount of time needed for image representation by contrast algorithm to be reduced. That work improved the image processing algorithm performance, and allowed the maximum usage of the multicores, by means of the Matlab.

## III. Parallel Image Processing

As mentioned, image processing can be integrated with parallel computing, such that image processing algorithms can be optimized from the time performance point of view, such a parallel program has features and decomposition type these aspects are covered in the following sub sections.

### 3.1 Parallel Program Features
Parallel processing software that is correct and efficient should consider the following features [9]:
1. Granularity: determines the level of program decomposition, such that one program is composed of some pieces, it is classified as:
a. Fine grained: many pieces but each has small size.
b. Coarse grained: few tasks but each has large size.
2. Parallelism Type:
a. Implicit: The compiler is the responsible for necessary instructions insertion to execute the program on a parallel computer.
b. Explicit: the program contains directives, written by the developer, they specify what processor does what part of job explicitly.
3. Synchronization: this is concerning how to coordinate between processors, to prevent tasks overlapping or deadlocks.
4. Network Latency: the delay measured from the send request up to the receive request, Network architecture plays an important role in such point.
5. Scalability: Parallel algorithm is said to be scalable if and only if program efficiency increases proportionally with increasing the number of processors.

### 3.2 Decomposition Classifications
Parallel processing could be classified into three types, Data, task and pipeline parallelism [13]:
1. Data Parallelism as in Fig.1, The same instruction runs on different blocks of data in the same time, such as the iterations of the parallel loop on an array, by these iterations must be independent of each other. This type of decomposition is known as "fine grained parallelism" because the amount of work is divided into many small portions. Alternatively, as Michael [12] said, "Data parallelism refers to any performer that has no dependences between iteration and the next."

**Figure 1:**Data parallelism

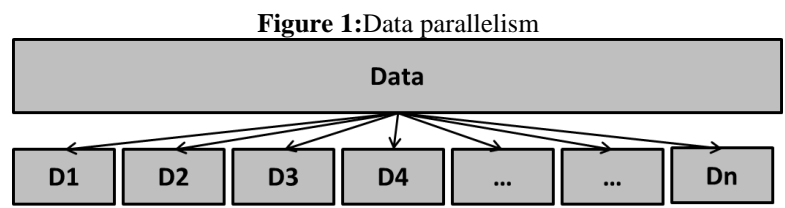

2. Task Parallelism as in Fig.2, Instead of the same operations being performed on different parts of the data, each process performs different operations. You can use task parallelism when your code can be divided into independent subroutines (methods/functions) that can be given to different processor and run simultaneously. This type of decomposition is known as "coarse grained parallelism" because the work is divided into few subtasks. More code is run in parallel because the parallelism is implemented at a higher level than in data parallelism. It is often easier to code and has less communication overhead than data parallelism [13]. As well, Michael [13] said, "Task parallelism refers to couple of performer that are on different parallel branches of the original stream graph."

**Figure 2:** Task parallelism

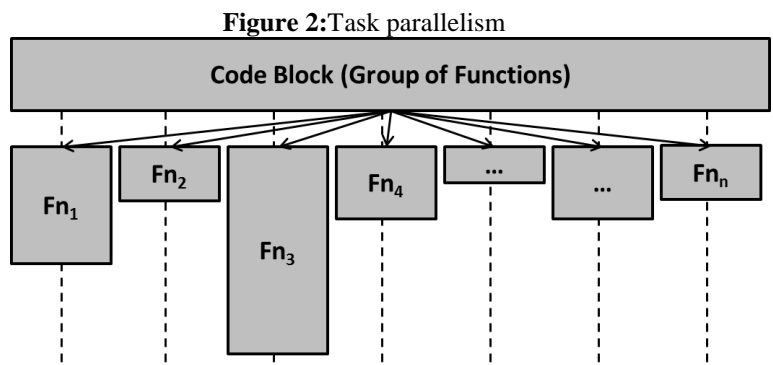

3. Pipeline Parallelism as in Fig.3, in this type of decomposition, long series of functions or tasks are in parallel, but there are some sort of overlapping in processes. it works as the output of one process could be the input of another process. In addition, Michael [13] expressed it as, "Pipeline parallelism applies to series of producers and consumers."

**Figure 3:** Pipeline parallelism

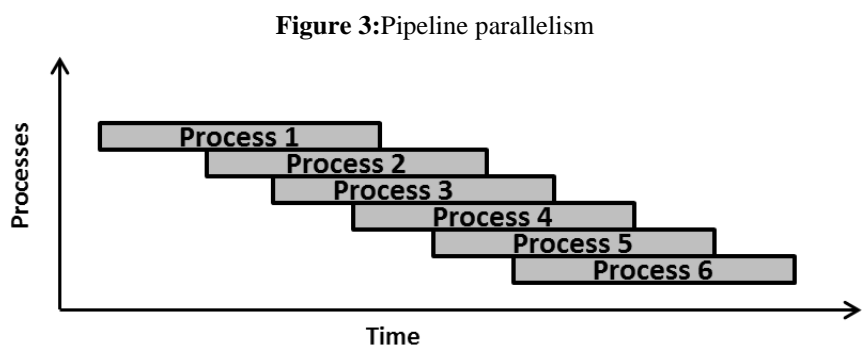

### 3.3 Parallel Processing Metrics
1. Speed up: Represented by Eq. 1. It is the ratio between serial and parallel execution time of a given algorithm [11]. If it is larger than one, this indicates that the parallel solution gives better results and vice versa.

$$S_P = \frac{T_s}{T_p} \quad (1)$$

Such that: $S_P$ is the speed up, $T_s$ is the sequential time and $T_p$ is the parallel time.

2. Efficiency: Eq.2 presents the efficiency of a parallel program and it is a measure of processor utilization, and calculated by the ratio of speed up and number of processors [11].

$$EFF = \frac{S_P}{P} \quad (2)$$

Such that: $EFF$ is the efficiency, $S_P$ is the speed up and $P$ is the number of processors.

3. Overheads of the parallel program are the extra times, required by the computations [11]. It is calculated by Eq. 3 calculates it.

$$T_{Over} = T_P - (T_S/P) \quad (3)$$

Such that $T_{Over}$ is the time of overheads, $T_P$ is the parallel time, $T_S$ is the serial time and $P$ is the number of processors.

*A Parallel Hybrid Technique for Multi-Noise Removal from Grayscale Medical Images*

**Figure 4:** Hybrid filter follow diagarm

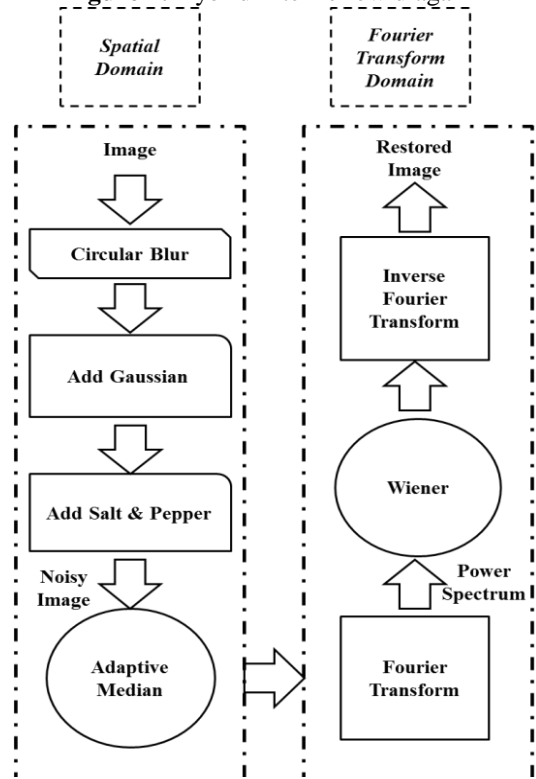

## IV. The Proposed Enhancement Methodology

Image enhancement and restoration in a noisy environment are common problems in image processing. Various filtering techniques have been developed for noise removal to improve the quality of images. Filters were designed assuming a specific noise distribution (Single-noise). For example, linear techniques are used to remove Gaussian noise as showed in [5], and order statistic techniques are used to remove impulsive i.e. salt & pepper noise. Hybrid filters have been developed to remove either Gaussian or impulsive noise (Multi-noise).

Fig.4 gives the architecture for the proposed hybrid approach in [6]. First, the adaptive median filter mechanism is applied in the spatial domain using MatLab standard libraries. In the other hand, the wiener filter is implemented in the frequency /Fourier transform domain. Additionally, Post processing operation is applied, Contrast stretching, after adaptive median filter and wiener filter, respectively. Adaptive median filter in spatial domain is done in convolution way, so it take longer time and it's a need to paralyze it in order to enhance time consumption using data parallelization architecture as represented in Fig.5.

**Figure 5:** (a) Hand medical image, (b) 2-partition version, (c) 4-partition version, (d) 8-partition version

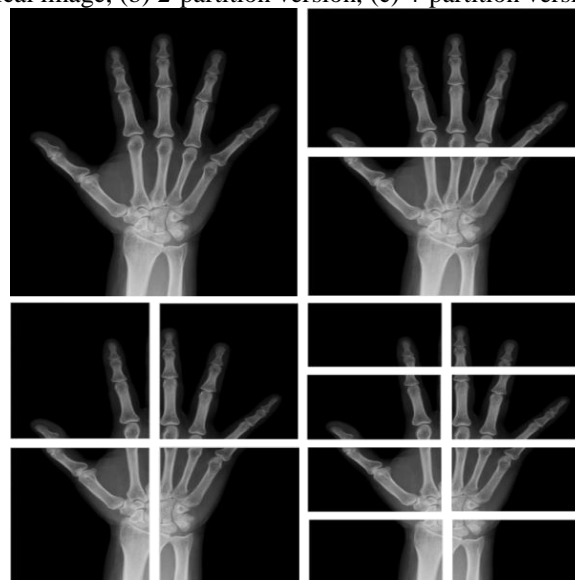

*A Parallel Hybrid Technique for Multi-Noise Removal from Grayscale Medical Images*

## 3.1 Parallel Hybrid Algorithm Implementation

Algorithm 1 illustrates parallel hybrid implementation. The implementation is basedonthree steps such that two core steps (a). apply parallel adaptive median and (b). Apply wiener filter on the output of the step (a) finally (c). a post processing step is applied for contrast stretching. The second procedure involves parallel implementation of adaptive median filter it takes the image, maximum window size and number of image parts or partitions. Briefly it apply a serial adaptive median filter but on all sub image in parallel using the parfor implicit parallelism powered by MatLab.

**Algorithm 1:** Parallel Hybrid Implementation Using MatLabParfor

```
1. original = imread(imPath);
2. original = im2double(original);
3. noisyBuf = degradImage(original);
4. procedure parallelHybrid (noisyBuf)
      a. tic();
      b. adapOnly = parallelAdapMedian(noisyBuf, 11, partition_i);
      c. partition_i belongs to {2, 4, 8};
      d. hybrid = wnrFn(adapOnly, original);
      e. hybrid = contrast(hybrid);
      f. fprintf('Time for Parallel Hybrid = %3.2f', toc(););
5. end procedure

6. procedure parallelAdapMedian(g, Smax, parts)
      a. if (Smax <= 1) | (Smax/2 == round(Smax/2)) | (Smax ~=
         round(Smax))
             i. error('SMAX must be an odd integer > 1.')
      b. end if
      c. [M, N] = size(g);
      d. f = g; %initial setup
      e. alreadyProcessed = false(size(g));
      f. splittedImg = imgSpliter( f, M, N, parts );
      g. parfor subimg = 1:parts, numberOfWorkers_i
             i. numberOfWorkers_i belongs to {2, 4, 6, 8, 10, 12}
            ii. myTemp = splittedImg(subimg);
           iii. t = alreadyProcessed;
            iv. for k = 3:2:Smax
                   1. zmin = getMinValue(splittedImg(subimg));
                   2. zmax = getMaxValue(splittedImg(subimg));
                   3. zmed = getMedianValue(splittedImg(subimg));
                   4. processByLvlB = (zmed > zmin) & (zmax > zmed) & ~t;
                   5. zB = (g > zmin) & (zmax > g);
                   6. outputZxy  = processByLvlB & zB;
                   7. outputZmed = processByLvlB & ~zB;
                   8. t = t | processByLvlB;
                   9. if( alreadyProcessed | processByLvlB)
                         a. if all(t(:))
                               i. continue;
                         b. end if
                  10. end if
                  11. myTemp(outputZxy) = g(outputZxy);
                  12. myTemp(outputZmed) = zmed(outputZmed);
             v. end for
            vi. splittedImg(subimg) = myTemp;
      h. end parfor
7. end procedure
```

**Table 1:** Time in seconds for serial hybrid and parallel hybrid when number of workers set to 2, 4, 6, 8 and 12 and image divided into 2, 4 and 8 partitions

| Image Dimensions | Serial | Workers | 2 | 4 | 6 | 8 | 10 | 12 |
|---|---|---|---|---|---|---|---|---|
| **1900 X 2368** | 79.5 | 2-Partitions | 10.41 | 10.87 | 10.89 | 10.37 | 10.96 | 11.11 |
| | | 4- Partitions | 12.1 | 11.99 | 12.22 | 12.29 | 12.49 | 12.23 |
| | | 8- Partitions | 16.22 | 15.84 | 15.87 | 16.62 | 15.64 | 16.2 |
| **3800 X 4736** | 272.42 | 2- Partitions | 44.67 | 41.59 | 36.14 | 35.12 | 42.86 | 49.28 |
| | | 4- Partitions | 41.37 | 40.76 | 40.62 | 41.06 | 41.35 | 42.18 |
| | | 8- Partitions | 54.88 | 54.65 | 56.63 | 56.85 | 54.69 | 55.21 |
| **7600 X 9472** | 1742.5 | 2- Partitions | 640.34 | 863.87 | 856.24 | 652.75 | 914.24 | 690.8 |
| | | 4- Partitions | 768 | 470.08 | 480.03 | 532.75 | 490.92 | 703.38 |
| | | 8- Partitions | 847.9 | 538.55 | 769.91 | 505.67 | 656.4 | 611.59 |

## V. Experimental Results & Discussion

The proposed hybrid approach was implemented in MatLab 2013a [6]. Parallel version of adaptive median filter is developed by using implicit MatLab parallelism – parfor, such that each MatLab workers can scatter the image among them, and then gather their local result to regenerate the image on the MatLab client. Fig.5 (a) shows the ground truth of input image (hand), (b) 2-partition input image, (c) shows 4-partition input image and (d) shows the 8-partition input image. Max window size for adaptive median filter is set to 11px that would require much time than 7px trial in [6].

### 5.1. Environment Settings
Our experiment specifications are as below; they are categorized into input, machine specs and tool used.
- Input:
    - ✓ Mode: Gray scale – no color info
    - ✓ Size: 277 KB, 715 KB and 2.5 MB
    - ✓ Dimensions: 1900 x 2368, 3800 x 4736 and 15200 x 18944
    - ✓ Resolution: 500px
- Machine specs:
    - ✓ Processor: Intel core i5
    - ✓ RAM: 6 GB
    - ✓ OS: Windows 7 enterprise edition, 64 bit
- Tool: MatLab 2013a,
    - ✓ Image Processing Toolbox
    Parallel Computing Toolbox

### 5.2 Results
Table 1 above gives a summary of our experiment such that, for each input (dimensions 1900 x2368, 3800 x 4736 and 7600 x 9472) the running time was measured for the sequential algorithm as well, for parallel hybrid algorithm number or workers and image partitions are varying. On one hand, workers number 2, 4, 6, 8, 10 and 12 workers have been tried. 12 workers setting is the max workers numbers allowed by local profile in MatLab. On the other hand, image partitions are 2, 4 and 8 as illustrated in Fig.5. Fig.6 gives the time consumed when images have been divided into 2, 4 and 8 partition, for each line the minimum time for 2-partition was when there are 2 threads, and for 4-partition image when there are 4 threads and for 8-partition image when threads are equal to 8. This is due to minimum communication and overhead as for each partition there is one and only one thread fully assigned to it.

**Figure 6:** Average time consumed in seconds for serial and parallel 2, 4 and 8 partition input.

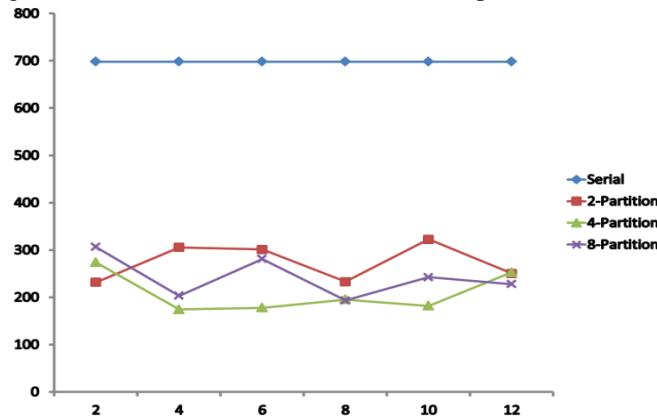

*A Parallel Hybrid Technique for Multi-Noise Removal from Grayscale Medical Images*

**Figure 7:** Average speed up for 2, 4 and 8 partition input

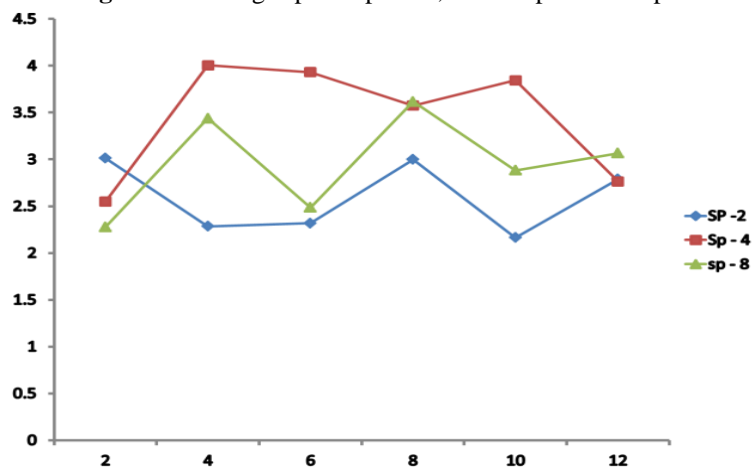

Fig.7 gives the resulted average speed up when image is divided into 2, 4 and 8 partitions the maximum speed up achieved in 4-partition case when assigned to 4 threads, it was = 4.05, 4-partition image assigned to 4 threads is the best practice for core i5 machine as there are only 4 cores associated into the process.

**Figure 8:** Efficiency for 2, 4 and 8 partition input

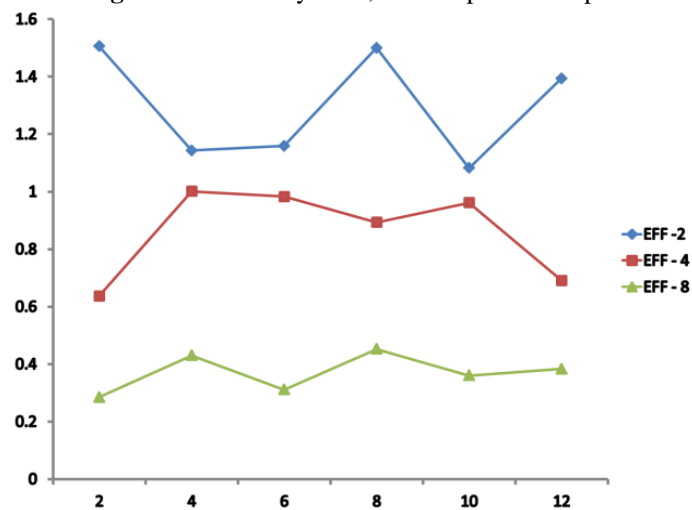

Fig.8 gives the efficiency chart for our average three image partition cases, it is clear that maximum efficiency for 2-partition image on 2 threads and it is equal to 1.51. However, for 4-partition image on 4 threads and it is equal to 1.01. Moreover for 8-partition image on eight threads and it is equal 0.45.

## VI. Conclusion

Medical images always have large sizes and they are commonly corrupted by single or multiple noise type at the same time, these two reasons were the triggers for moving toward parallel image processing to find alternatives techniques for image de-noising. This paper presented a parallel hybrid (adaptive median and wiener) filter implementation for gray scale medical image de-noising, based on a parallelization for adaptive median filter. The implementation was tested on gray scale medical image of 277 KB, 715 KB and 2.5 MB size, which each was divided into 2, 4 and 8 partitions; the proposed implementation was compared to the sequential implementation in terms of time. Thus, each case has the best time when assigned to number of threads equal to the number of its partitions. Such that, 2-partition image consumed average minimum time 231.81 seconds when number of threads was 2, 4-partition consumed average minimum time 174.28 seconds when number of threads was 4 and 8-partition consumed average minimum time 193.05 seconds when the number of threads was 8. In addition, Parallel speed up was calculated, and gave up to 3.01, 4.01 and 3.62 for 2, 4 and 8 partition consequently. Moreover, Efficiency was calculated, and gave up to 1.51, 1.01 and 0.45 for 2, 4 and 8 partitions cases in sequence.

DOI: 10.9790/0661-180505121128                                      www.iosrjournals.org                                                    127 | Page

## References

[1]. Salem S. Al-amri, Dr. N.V. Kalyankar, and Dr. Khamitkar S.D, A comparative study of removal noise from remote sensing image, *IJCSI, Vol. 7, No. 1*, 2010, 32 - 35.
[2]. Muhammad Sharif, AyyazHussain, Muhammad ArfanJaffar and Tae-Sun Choi, Fuzzy-based hybrid filter for Rician noise removal, *J. Signal, Image and Video Processing, Springer, London*, 2015, 1-10.
[3]. Ankita D. and Archana S., An advanced filter for image enhancement and restoration, *OJAET, Vol.1, No. 1*, 2013, 7-10.
[4]. Zhou W. Alan C. Bovik, Hamid R. Sheikh and Eero P. Simoncelli, Image quality assessment: from error visibility to structural similarity, *IEEE Transactions On Image Processing, Vol. 13, No. 4*, 2004, 600 - 605.
[5]. Nora Youssef, Abeer M. Mahmoud and El-Sayed M. El-Horbaty,Gaussian de-noising techniques in spatial domain for gray scale medical images, *ITK, Vol. 8, No.3*, 2014, 90-100.
[6]. Nora Youssef, Abeer M. Mahmoud and El-Sayed M. El-Horbaty, A hybrid de-nosing technique for multi-noise removal on gray scale medical images, *IJTS, Vol. 28, No.2*, 2015, 06-116.
[7]. Seema and Meenakshi G., Wavelet based technique for removal of multiple noises simultaneously, *IJACEN, Vol.2, No.1*, 2014, 34-38.
[8]. Shreyamsha Kumar, B.K. Image de-noising based on non-local means filter and its method noise thresholding, *J Signal, Image and Video Processing, Springer London, Vol.7, No.6*, 2013, 1211-1227.
[9]. Sanjay S. Neeraj Sharma and ShiruSharma,Image processing tasks using parallel computing in multi core architecture and its applications in medical imaging, *IJARCCE, Vol. 2, No. 4*, 2013, 1896-1899.
[10]. Sharanjit Singh Parneetkaur and Kamaldeepkaur, Parallel computing in digital image processing, *IJARCCE, Vol. 4, No. 1*, 2015, 183-186.
[11]. Er.Paramjeetkaur and Er.Nishi: A survey on CUDA, *IJCSIT, Vol.5 No.2*, 2014, 2210-2214.
[12]. Preetikaur, Implementation of image processing algorithms on the parallel platform using MatLab, *IJCSET, Vol. 4, No. 06*, 2013, 696-706.
[13]. AnanthGrama, Anshul Gupta, George Karypis and Vipin Kumar, Introduction to parallel computing 2[nd] Ed. (Addison Wesly, 2003).
[14]. Michael I. Gordon, William Thies and SamanAmarasinghe, Exploiting coarse-grained task, data, and pipeline parallelism in stream programs, *Proc. 12[th] ASPLOS XII Int. Conf. Architectural support for programming languages and operating systems*, California, USA, 2006, 151-162.

**Nora Youssef:** is a research assistant at the university of Ain Shams, Egypt. where she is currently purses her M.Sc. degree in Computer and information Science, Her research interests are digital image enhancement and restoration and parallel computing, her current work focus is on building a parallel hybrid algorithm based on adaptive median and wiener filters for Gaussian, Salt & Pepper and blurriness removal in gray scale medical images domain.

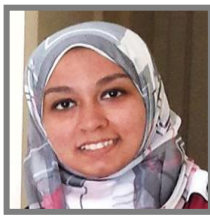

**Abeer M. Mahmoud:** She received her Ph.D. (2010) in Computer science from Niigata University, Japan, her M.Sc (2004) B.Sc. (2000) in computer science from Ain Shams University, Egypt. Her work experience is as a lecturer assistant and assistant professor, faculty, of computer and information sciences, Ain Shams University. Her research areas include artificial intelligence medical data mining, machine learning, and robotic simulation systems.

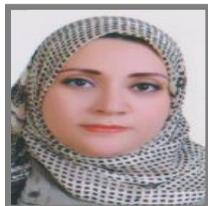

**El-Sayed M. El-Horbaty**: He received his Ph.D. in Computer science from London University, U.K., his M.Sc. (1978) and B.Sc (1974) in Mathematics FromAin Shams University, Egypt. His work experience includes 39 years as an in Egypt (Ain Shams University), Qatar (Qatar University) and Emirates (Emirates University, Ajman University and ADU University). He Worked as Deputy Dean of the faculty of IT, Ajman University (2002-2008). He is working as a Vice Dean of the faculty of Computer & Information Sciences, Ain Shams University (2010-Now). Prof. El-Horbaty is current areas of research are parallel algorithms, combinatorial optimization, image processing. His work appeared in journals such as Parallel Computing, International journal of Computers and Applications (IJCA), Applied Mathematics and Computation, and International Review on Computers and software. In addition, he has been involved in more than 26 conferences.

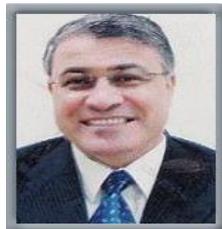